\documentclass[11pt,twocolumn]{article}
\usepackage{amsmath,graphicx}
\usepackage{dsfont}
\usepackage{subcaption}

\title{Learning Sparse Wavelet Representations}

\author{Daniel Recoskie\thanks{Thanks to NSERC for funding.}~~and~Richard Mann\\
	University of Waterloo\\
	David R. Cheriton School of Computer Science\\
	Waterloo, Canada\\
	\{dprecosk, mannr\}@uwaterloo.ca}
\date{}

\begin{document}
\maketitle
\begin{abstract}
In this work we propose a method for learning wavelet filters directly from data. We accomplish this by framing the discrete wavelet transform as a modified convolutional neural network. We introduce an autoencoder wavelet transform network that is trained using gradient descent. We show that the model is capable of learning structured wavelet filters from synthetic and real data. The learned wavelets are shown to be similar to traditional wavelets that are derived using Fourier methods. Our method is simple to implement and easily incorporated into neural network architectures. A major advantage to our model is that we can learn from raw audio data.
\end{abstract}
%
%
\section{Introduction}
\label{sec:intro}
The wavelet transform has several useful properties that make it a good choice for a feature representation including: a linear time algorithm, perfect reconstruction, and the ability to tailor wavelet functions to the application. However, the wavelet transform is not widely used in the machine learning community. Instead, methods like the Fourier transform and its variants are often used (e.g. \cite{graves2013speech}). We believe that one cause of the lack of use is the difficulty in designing and selecting appropriate wavelet functions. Wavelet filters are typically derived analytically using Fourier methods. Furthermore, there are many different wavelet functions to choose from. Without a deep understanding of wavelet theory, it can be difficult to know which wavelet to choose. This difficulty may lead many to stick to simpler methods.

We propose a method that learns wavelet functions directly from data using a neural network framework. As such, we can leverage the theoretical properties of the wavelet transform without the difficult task of designing or choosing a wavelet function. An advantage of this method is that we are able to learn directly from raw audio data. Learning from raw audio has shown success in audio generation \cite{oord2016wavenet}.

We are not the first to propose using wavelets in neural network architectures. There has been previous work in using fixed wavelet filters in neural networks such as the wavelet network \cite{zhang1992wavelet} and the scattering transform \cite{mallat2012group}. Unlike our proposed method, these works do not learn wavelet functions from data. 

One notable work involving learning wavelets can be found in \cite{waveletGraphs}. Though the authors also propose learning wavelets from data, there are several differences from our work. One major difference is that second generation wavelets are considered instead of the traditional (first generation) wavelets considered here \cite{sweldens1998lifting}. Secondly, the domain of the signals were over the vertices of graphs, as opposed to $\mathds{R}$. 

We begin our discussion with the wavelet transform. We will provide some mathematical background as well as outline the discrete wavelet transform algorithm. Next, we outline our proposed wavelet transform model. We show that we can represent the wavelet transform as a modified convolutional neural network. We then evaluate our model by demonstrating we can learn useful wavelet functions by using an architecture similar to traditional autoencoders \cite{hinton2006reducing}. 

\section{Wavelet transform}
We choose to focus on a specific type of linear time-frequency transform known as the {\em wavelet transform}. The wavelet transform makes use of a dictionary of wavelet functions that are dilated and shifted versions of a mother wavelet. The mother wavelet, $\psi$, is constrained to have zero mean and unit norm. The dilated and scaled wavelet functions are of the form:
\begin{equation}
\psi_j[n] = \frac{1}{2^j} \psi \left( \frac{n}{2^j} \right).
\end{equation}
where $n,j \in \mathds{Z}$. The discrete wavelet transform is defined as
\begin{equation}
\label{eq:wt}
W x[n, 2^j] = \sum_{m=0}^{N-1} x[m] \psi^*_j[m-n]
\end{equation}
for a discrete real signal $x$.

The wavelet functions can be thought of as a bandpass filter bank. The wavelet transform is then a decomposition of a signal with this filter bank. Since the wavelets are bandpass, we require the notion of a lowpass scaling function that is the sum of all wavelets above a certain scale $j$ in order to fully represent the signal. We define the scaling function, $\phi$, such that its Fourier transform, $\hat{\phi}$, satisfies
\begin{equation}
|\hat{\phi}(\omega)|^2 = \int^{+\infty}_1 \frac{|\hat{\psi}(s\omega)|^2}{s} ds
\end{equation}
with the phase of $\hat{\phi}$ being arbitrary \cite{mallat}.

The discrete wavelet transform and its inverse can be computed via a fast decimating algorithm. Let us define two filters

\begin{equation}
h[n] = \left\langle \frac{1}{\sqrt{2}} \phi\left(\frac{t}{2}\right), \phi(t-n) \right\rangle
\label{eq:wavelet-filter-coefficients}
\end{equation}

\begin{equation}
g[n] =\left\langle  \frac{1}{\sqrt{2}} \psi\left(\frac{t}{2}\right), \phi(t-n) \right\rangle
\label{eq:scaling-filter-coefficients}
\end{equation}

The following equations connect the wavelet coefficients to the filters $h$ and $g$, and give rise to a recursive algorithm for computing the wavelet transform.

\noindent Wavelet Filter Bank Decomposition:
\begin{equation}
\label{eq:dwta}
a_{j+1}[p] = \sum^{+\infty}_{n=-\infty} h[n-2p]a_j[n]
\end{equation}
\begin{equation}
\label{eq:dwtd}
d_{j+1}[p] = \sum^{+\infty}_{n=-\infty} g[n-2p]a_j[n]
\end{equation}

\noindent Wavelet Filter Bank Reconstruction
\begin{equation}
\label{eq:swt}
a_{j}[p] = \sum^{+\infty}_{n=-\infty} h[p-2n]a_{j+1}[n] + \sum^{+\infty}_{n=-\infty} g[p-2n]d_{j+1}[n]
\end{equation}

We call $a$ and $d$ the approximation and detail coefficients respectively. The detail coefficients are exactly the wavelet coefficients defined by Equation \ref{eq:wt}. As shown in Equations \ref{eq:dwta} and \ref{eq:dwtd}, the wavelet coefficients are computed by recursively computing the coefficients at each scale, with $a_0$ initialized with the signal $x$. At each step of the algorithm, the signal is split into high and low frequency components by convolving the approximation coefficients with $h$ (scaling filter) and $g$ (wavelet filter).  The low frequency component becomes the input to the next step of the algorithm. Note that $a_i$ and $d_i$ are downsampled by a factor of two at each iteration. An advantage of this algorithm is that we only require two filters instead of an entire filter bank. The wavelet transform effectively partitions the signal into frequency bands defined by the wavelet functions. 
We can reconstruct a signal from its wavelet coefficients using Equation \ref{eq:swt}. We call the reconstruction algorithm the inverse discrete wavelet transform. A thorough treatment of the wavelet transform can be found in \cite{mallat}.

\begin{figure}[!htb]
\begin{minipage}[b]{1.0\linewidth}
  \centering
\includegraphics[width=.95\textwidth]{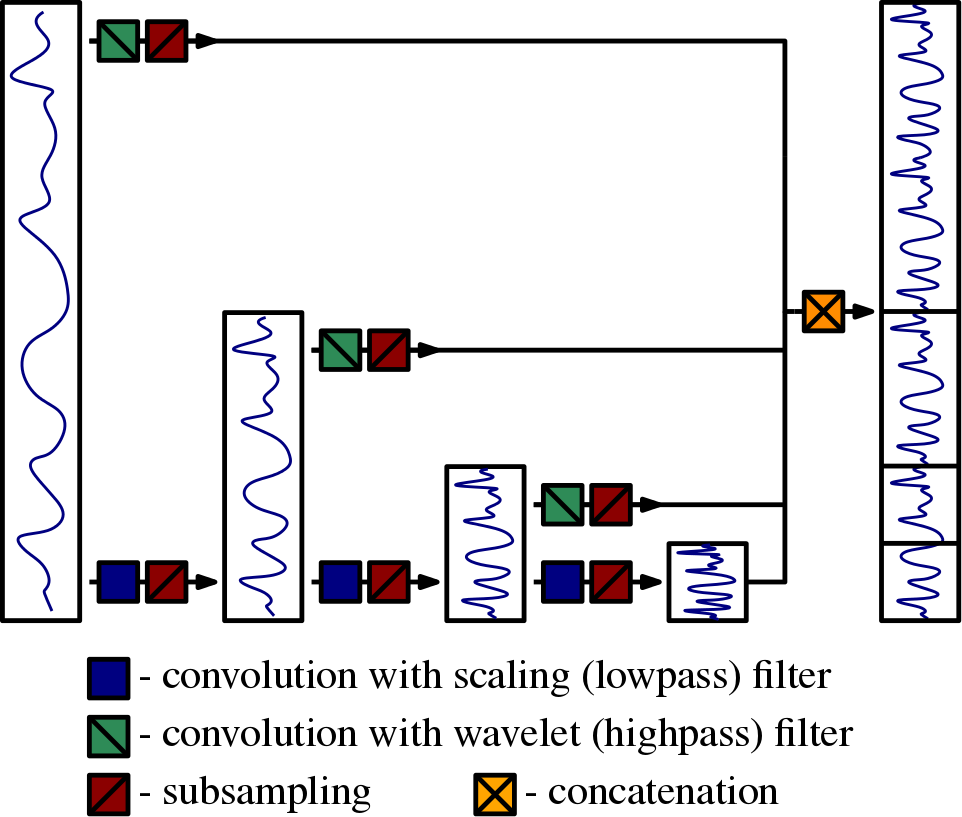}
\end{minipage}
\caption{The discrete wavelet transform represented as a neural network. This network computes the discrete wavelet transform of its input using Equations \ref{eq:dwta} and \ref{eq:dwtd}}
\label{fig:wnn}
\end{figure}

\section{Proposed Model}
We propose a method for learning wavelet functions by defining the discrete wavelet transform as a convolutional neural network (CNN). CNNs compute a feature representation of an input signal through a cascade of filters. They have seen success in many signal processing tasks, such as speech recognition and music classification \cite{sainath2013deep, choi2017convolutional}. Generally, CNNs are not applied directly to raw audio data. Instead, a transform is first applied to the signal (such as the short-time Fourier transform). This representation is then fed into the network. 

Our proposed method works directly on the raw audio signal. We accomplish this by implementing the discrete wavelet transform as a modified CNN. Figure \ref{fig:wnn} shows a graphical representation of our model, which consists of repeated applications of Equations \ref{eq:dwta} and \ref{eq:dwtd}. The parameters (or weights) of this network are the wavelet and scaling filters $g$ and $h$. Thus, the network computes the wavelet coefficients of a signal, but allows the wavelet filter to be learned from the data. We can similarly define an inverse network using Equation \ref{eq:swt}.

We can view our network as an unrolling of the discrete wavelet transform algorithm similar to unrolling a recurrent neural network (RNN) \cite{pascanu2013difficulty}. Unlike an RNN, our model takes as input the entire input signal and reduces the scale at every layer through downsampling. Each layer of the network corresponds to one iteration of the algorithm. At each layer, the detail coefficients are passed directly to the final layer. The final layer output, denoted $W(x)$, is formed as a concatenation of all the computed detail coefficients and the final approximation coefficients. We propose that this network be used as an initial module as part of a larger neural network architecture. This would allow a neural network architecture to take as input raw audio data, as opposed to some transformed version.

We restrict ourselves to quadrature mirror filters. That is, we set
\begin{equation}
g[n] = (-1)^n h[-n]
\end{equation}
By making this restriction, we reduce our parameters to only the scaling filter $h$. 

\begin{figure}[tb]
\begin{minipage}[b]{1.0\linewidth}
\centering
\includegraphics[width=.45\textwidth]{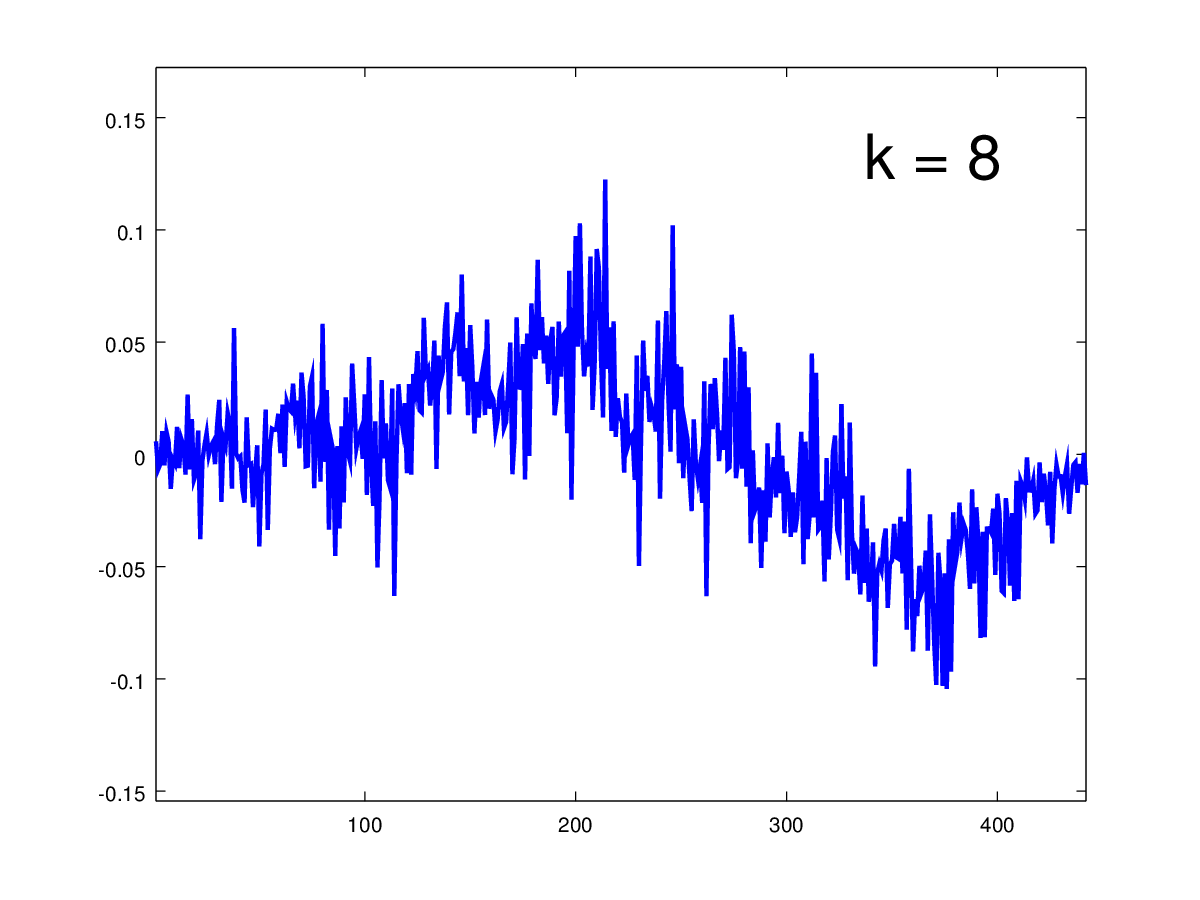}
\includegraphics[width=.45\textwidth]{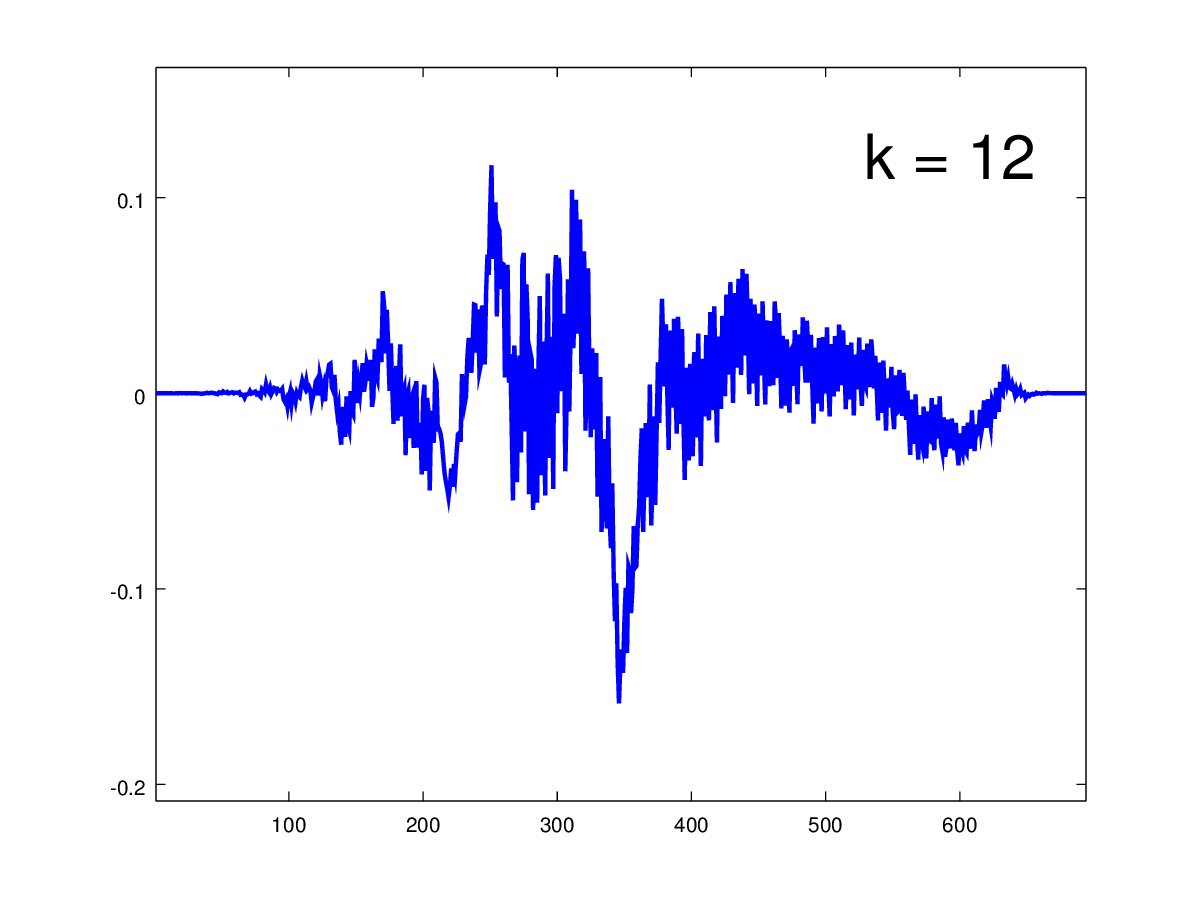}
\includegraphics[width=.45\textwidth]{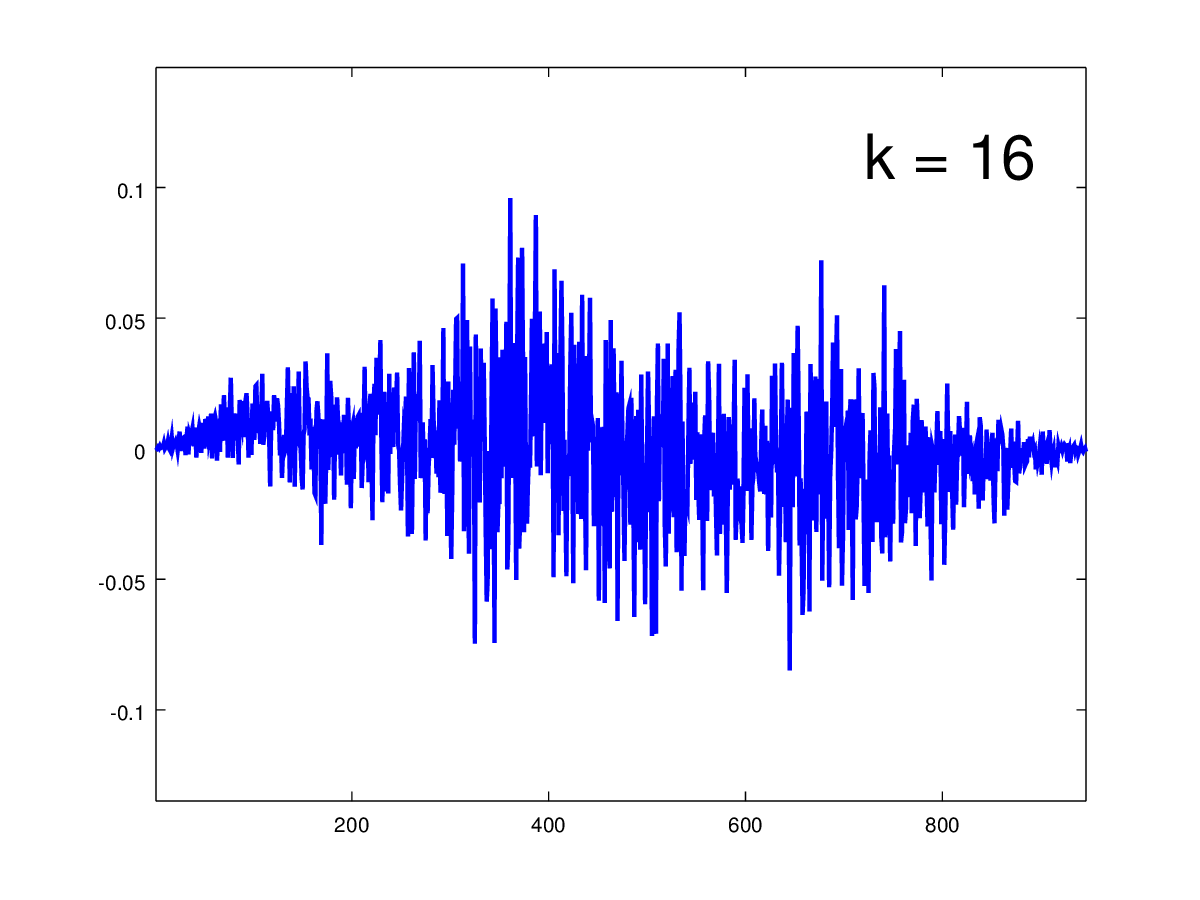}
\includegraphics[width=.45\textwidth]{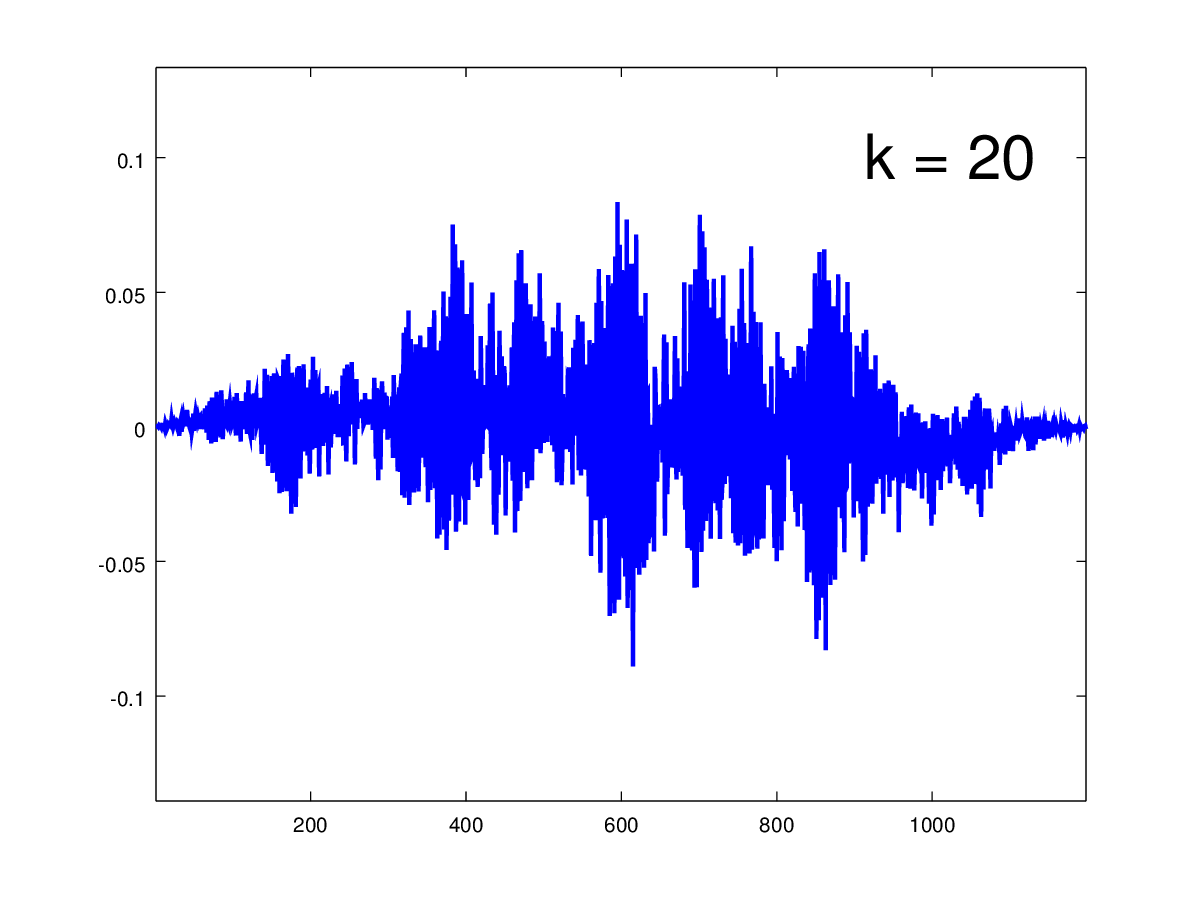}
\end{minipage}
\caption{Examples of random wavelet functions that satisfy Equation \ref{eq:wavelet_constraints} for different filter lengths.}
\label{fig:random_wavelets}
\end{figure}

The model parameters will be learned by gradient descent. As such, we must introduce constraints that will guarantee the model learns wavelet filters. We define the wavelet constraints as
\begin{equation}
L_w(h,g) = (||h||_2 - 1)^2 + (\mu_h - \sqrt{2}/k)^2 + \mu_g^2
\label{eq:wavelet_constraints}
\end{equation}
where $\mu_h$ and $\mu_g$ are the means of $h$ and $g$ respectively, and $k$ is length of the filters. The first two terms correspond to finite $L_2$ and $L_1$ norms respectively. The third term is a relaxed orthogonality constraint. Note that these are soft constraints, and thus the filters learned by the model are only approximately wavelet filters. See Figure \ref{fig:random_wavelets} for examples of randomly chosen wavelet functions derived from filters that minimize Equation \ref{eq:wavelet_constraints}. We have not explored the connection between the space of wavelets that minimize Equation \ref{eq:wavelet_constraints} and those of parameterized wavelet families \cite{burrus1997introduction}.

\section{Evaluation}
\begin{figure}[tb]
\begin{minipage}[b]{1.0\linewidth}
\includegraphics[width=1\textwidth]{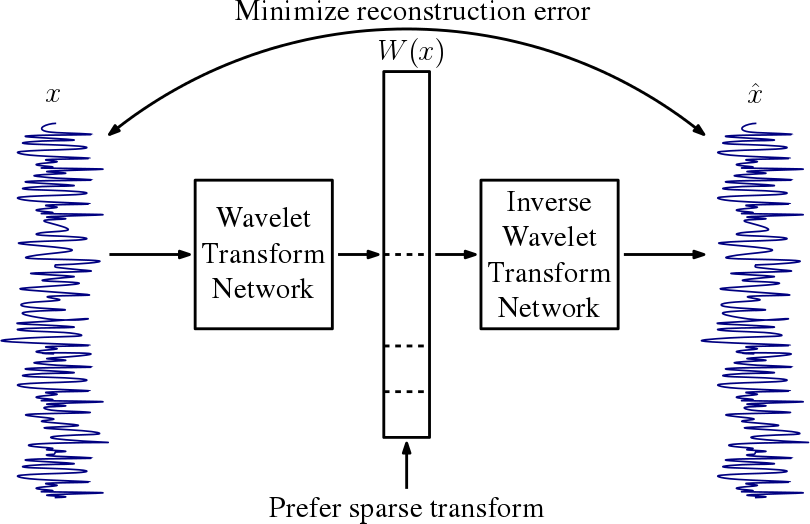}
\end{minipage}
\caption{The reconstruction network is composed of a wavelet transform followed by an inverse wavelet transform.}
\label{fig:wnae}
\end{figure}

We will evaluate our wavelet model by learning wavelet filters that give sparse representations. We achieve this by constructing an autoencoder as illustrated in Figure \ref{fig:wnae}. Autoencoders are used in unsupervised learning in order to learn useful data representations \cite{hinton2006reducing}. Our autoencoder is composed of a wavelet transform network followed by an inverse wavelet transform network. The loss function is made up of a reconstruction loss, a sparsity term, and the wavelet constraints. Let $\hat{x}_i$ denote the reconstructed signal. The loss function is defined as
\begin{equation}
\begin{split}
L(X; g,h) =& \frac{1}{M}\sum_{i=1}^{M} ||x_i - \hat{x}_i||_2^2 \\
	+& \lambda_1 \frac{1}{M}\sum_{i=1}^{M} ||W(x_i)||_1 + \lambda_2 L_w(h,g)
\end{split}
\label{eq:loss}
\end{equation}
for a dataset $X = \{x_1, x_2, \ldots, x_M\}$ of fixed length signals. In our experiments, we fix $\lambda_1 = \lambda_2 = 1/2$.

We conducted experiments on synthetic and real data. The real data consists of segments taken from the MIDI aligned piano dataset (MAPS) \cite{emiya2010multipitch}. The synthetic data consists of harmonic data generated from simple periodic waves. We construct a synthetic signal, $x_i$, from a base periodic wave function, $s$, as follows:
\begin{equation}
x_i(t) = \sum_{k=0}^{K-1} a_k \cdot s(2^k t + \phi_k)
\label{eq:synth-harmonic}
\end{equation}
where $\phi_k \in [0, 2\pi]$ is a phase offset chosen uniformly at random, and $a_k \in \{0, 1\}$ is the $k^{th}$ harmonic indicator which takes the value of 1 with probability $p$. We considered three different base waves: sine, sawtooth, and square. A second type of synthetic signal was created similarly to Equation \ref{eq:synth-harmonic} by windowing the base wave at each scale with randomly centered Gaussian windows (multiple windows at each scale were allowed). The length of the learned filters is 20 in all trials. The length of each $x_i$ is 1024 and we set $K=5$, $p=1/2$, and $M=32000$. We implemented our model using Google's Tensorflow library \cite{tensorflow-short}. We make use of the Adam algorithm for stochastic gradient descent \cite{kingma2014adam}. We use a batch size of 32 and run the optimizer until convergence.

\begin{figure}[tb]
\begin{minipage}[b]{1.0\linewidth}
\centering
 \includegraphics[width=.32\textwidth]{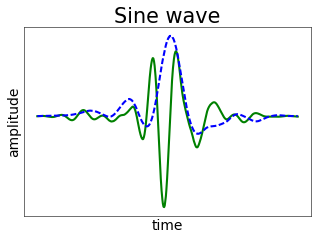}
 \includegraphics[width=.32\textwidth]{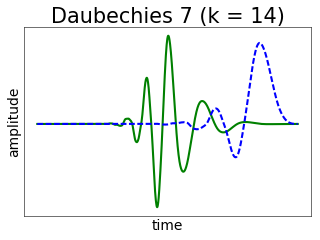}
 \includegraphics[width=.32\textwidth]{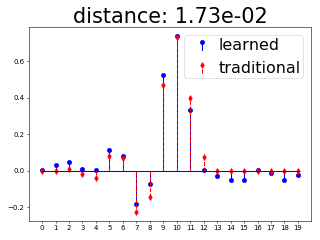}
 \includegraphics[width=.32\textwidth]{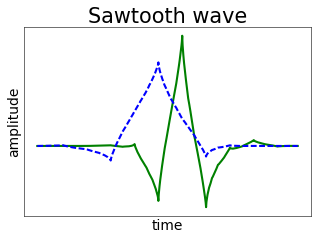}
\includegraphics[width=.32\textwidth]{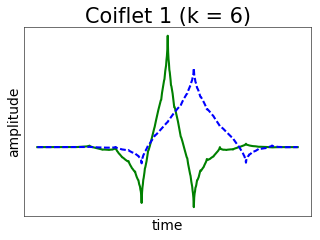}
\includegraphics[width=.32\textwidth]{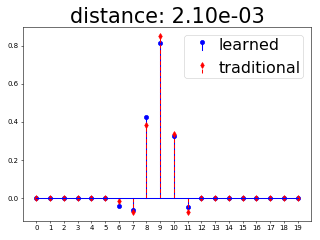}
 \includegraphics[width=.32\textwidth]{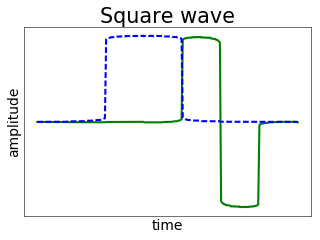}
\includegraphics[width=.32\textwidth]{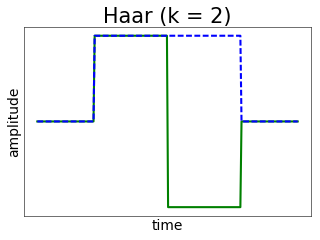}
\includegraphics[width=.32\textwidth]{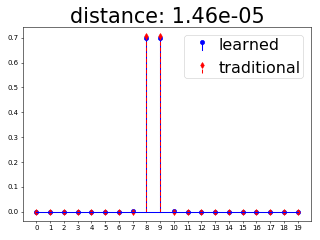}
 \includegraphics[width=.32\textwidth]{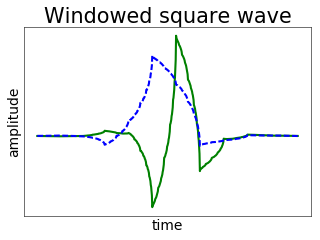}
\includegraphics[width=.32\textwidth]{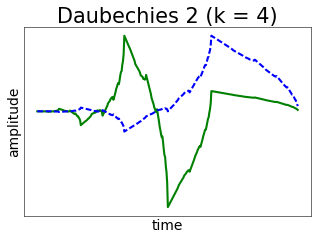}
\includegraphics[width=.32\textwidth]{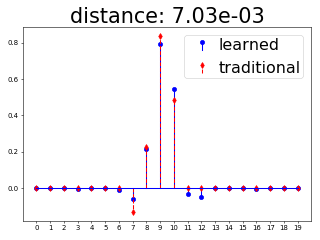}
\includegraphics[width=.32\textwidth]{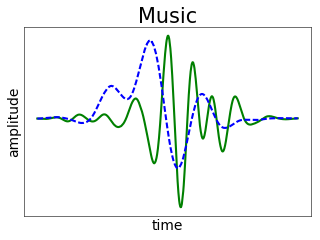}
\includegraphics[width=.32\textwidth]{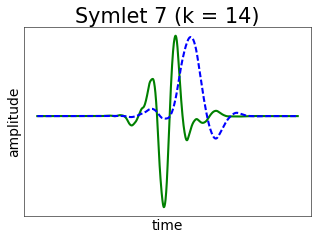}
\includegraphics[width=.32\textwidth]{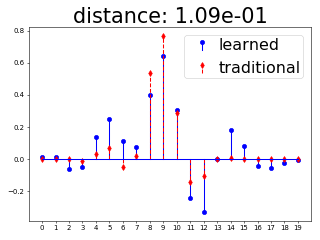}
\end{minipage}
\caption{Left column: Learned  wavelet (solid) and scaling (dashed) functions. Middle column: Closest traditional wavelet (solid) and scaling (dashed) functions. Right column: Plots of the scaling filters from the first two columns with corresponding distance measure.}
\label{fig:results}
\end{figure}

The wavelet filters learned are unique to the type of data that is being reconstructed. Example wavelet functions are included in Figure \ref{fig:results}. These functions are computed from the scaling filter coefficients using the cascade algorithm \cite{strang1996wavelets}. Note that the learned functions are highly structured, unlike the random wavelet functions in Figure \ref{fig:random_wavelets}. 

In order to compare the learned wavelets to traditional wavelets, we will first define a distance measure between filters of length $k$:
\begin{equation}
dist(h_1, h_2) = \min_{0 \le i < k} 1 - \frac{\langle h_1, \text{shift}(h_2,i) \rangle}{||h_1||_2 \cdot ||h_2||_2}
\label{eq:dist}
\end{equation}
where $\text{shift}(h_2,i)$ is $h_2$ circular shifted by $i$ samples. This measure is the minimum cosine distance under all circular shifts of the filters. To compare different length filters, we zero-pad the shorter filter to the length of the longer. We restrict our consideration to the following traditional wavelet families: Haar, Daubechies, Symlets, and Coiflets. The middle column of Figure \ref{fig:results} shows the closest traditional wavelet to the learned wavelets according to Equation \ref{eq:dist}. The distances are listed in the right column.

\begin{figure}[tb]
\begin{minipage}[b]{1.0\linewidth}
\centering
\includegraphics[width=.45\textwidth]{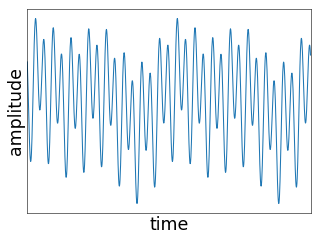}
\includegraphics[width=.45\textwidth]{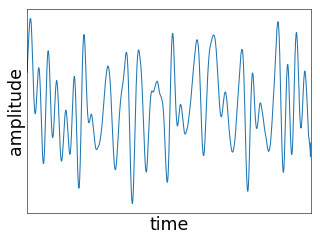}
\\
\includegraphics[width=.45\textwidth]{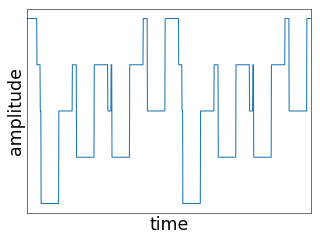}
\includegraphics[width=.45\textwidth]{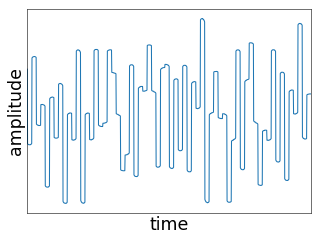}
\\
\includegraphics[width=.45\textwidth]{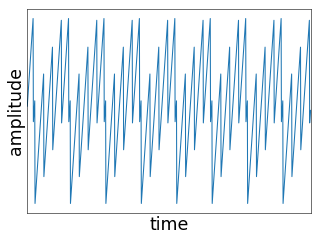}
\includegraphics[width=.45\textwidth]{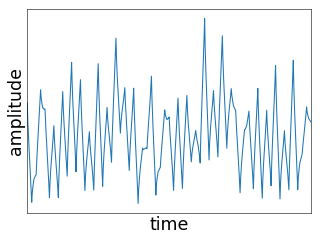}
\end{minipage}
\caption{Left column: Examples of synthetic training signals. Right column: Examples of signals generated from the corresponding learned wavelet filters. Base waves from top to bottom: sine, square, sawtooth.}
\label{fig:gen_wavs}
\end{figure}

In order to determine how well the learned wavelets capture the structure of the training data signals, we will consider signals randomly generated from the learned wavelets. To generate signals we begin by sparsely populating wavelet coefficients from $[-1, 1]$. The coefficients from the three highest frequency scales are then set to zero. Finally, the generated signal is obtained by performing an inverse wavelet transform of the sparse coefficients. Qualitative results are shown in Figure \ref{fig:gen_wavs}. Typical training examples are shown in the left column. Example generated signals are shown in the right column. Note that the generated signals have visually similar structure to the training examples. This provides evidence that the learned wavelets have captured the structure of the data.

\section{Conclusion}
We have proposed a new model capable of learning useful wavelet representations from data. We accomplish this by framing the wavelet transform as a modified CNN. We show that we can learn useful wavelet filters by gradient descent, as opposed to the traditional derivation of wavelets using Fourier methods. The learned wavelets are able to capture the structure of the data. We hope that our work leads to wider use of the wavelet transform in the machine learning community.

Framing our model as a neural network has the benefit of allowing us to leverage deep learning software frameworks, and also allows for simple integration into existing neural network architectures. An advantage of our method is the ability to learn directly from raw audio data, instead of relying on a fixed representation such as the Fourier transform.

\bibliographystyle{plain}
{\small \bibliography{}}

\begin{thebibliography}{10}

\bibitem{tensorflow-short}
Mart\'{\i}n Abadi et~al.
\newblock {TensorFlow}: Large-scale machine learning on heterogeneous systems,
  2015.
\newblock Software available from tensorflow.org.

\bibitem{burrus1997introduction}
C~Sidney Burrus, Ramesh~A Gopinath, and Haitao Guo.
\newblock {\em Introduction to wavelets and wavelet transforms: a primer}.
\newblock Prentice-Hall, Inc., 1997.

\bibitem{choi2017convolutional}
Keunwoo Choi, Gy{\"o}rgy Fazekas, Mark Sandler, and Kyunghyun Cho.
\newblock Convolutional recurrent neural networks for music classification.
\newblock In {\em Acoustics, Speech and Signal Processing (ICASSP), 2017 IEEE
  International Conference on}, pages 2392--2396. IEEE, 2017.

\bibitem{emiya2010multipitch}
Valentin Emiya, Roland Badeau, and Bertrand David.
\newblock Multipitch estimation of piano sounds using a new probabilistic
  spectral smoothness principle.
\newblock {\em IEEE Transactions on Audio, Speech, and Language Processing},
  18(6):1643--1654, 2010.

\bibitem{graves2013speech}
Alex Graves, Abdel-rahman Mohamed, and Geoffrey Hinton.
\newblock Speech recognition with deep recurrent neural networks.
\newblock In {\em Acoustics, speech and signal processing (icassp), 2013 ieee
  international conference on}, pages 6645--6649. IEEE, 2013.

\bibitem{hinton2006reducing}
Geoffrey~E Hinton and Ruslan~R Salakhutdinov.
\newblock Reducing the dimensionality of data with neural networks.
\newblock {\em science}, 313(5786):504--507, 2006.

\bibitem{kingma2014adam}
Diederik Kingma and Jimmy Ba.
\newblock Adam: A method for stochastic optimization.
\newblock {\em arXiv preprint arXiv:1412.6980}, 2014.

\bibitem{mallat}
St{\'e}phane Mallat.
\newblock {\em A Wavelet Tour of Signal Processing, Third Edition: The Sparse
  Way}.
\newblock Academic Press, 3rd edition, 2008.

\bibitem{mallat2012group}
St{\'e}phane Mallat.
\newblock Group invariant scattering.
\newblock {\em Communications on Pure and Applied Mathematics},
  65(10):1331--1398, 2012.

\bibitem{oord2016wavenet}
Aaron van~den Oord, Sander Dieleman, Heiga Zen, Karen Simonyan, Oriol Vinyals,
  Alex Graves, Nal Kalchbrenner, Andrew Senior, and Koray Kavukcuoglu.
\newblock Wavenet: A generative model for raw audio.
\newblock {\em arXiv preprint arXiv:1609.03499}, 2016.

\bibitem{pascanu2013difficulty}
Razvan Pascanu, Tomas Mikolov, and Yoshua Bengio.
\newblock On the difficulty of training recurrent neural networks.
\newblock In {\em International Conference on Machine Learning}, pages
  1310--1318, 2013.

\bibitem{waveletGraphs}
Raif Rustamov and Leonidas~J Guibas.
\newblock Wavelets on graphs via deep learning.
\newblock In {\em Advances in Neural Information Processing Systems}, pages
  998--1006, 2013.

\bibitem{sainath2013deep}
Tara~N Sainath, Abdel-rahman Mohamed, Brian Kingsbury, and Bhuvana Ramabhadran.
\newblock Deep convolutional neural networks for lvcsr.
\newblock In {\em Acoustics, Speech and Signal Processing (ICASSP), 2013 IEEE
  International Conference on}, pages 8614--8618. IEEE, 2013.

\bibitem{strang1996wavelets}
Gilbert Strang and Truong Nguyen.
\newblock {\em Wavelets and filter banks}.
\newblock SIAM, 1996.

\bibitem{sweldens1998lifting}
Wim Sweldens.
\newblock The lifting scheme: A construction of second generation wavelets.
\newblock {\em SIAM journal on mathematical analysis}, 29(2):511--546, 1998.

\bibitem{zhang1992wavelet}
Qinghua Zhang and Albert Benveniste.
\newblock Wavelet networks.
\newblock {\em IEEE transactions on Neural Networks}, 3(6):889--898, 1992.

\end{thebibliography}

\end{document}